\documentclass{article}

\usepackage[english]{babel}
\usepackage[letterpaper,top=2cm,bottom=2cm,left=3cm,right=3cm,marginparwidth=1.75cm]{geometry}

\usepackage{amsmath}

\usepackage{subfig}
\usepackage{graphicx}
\usepackage{amssymb}
\usepackage{booktabs}
\usepackage{array}
\usepackage{enumitem}
\usepackage{placeins}
\usepackage[colorlinks=true, allcolors=blue]{hyperref}
\usepackage{xcolor}
\usepackage{pifont}

\usepackage{tikz}
\usetikzlibrary{shapes.geometric, arrows.meta, positioning, fit, backgrounds}

\tikzset{
  phase/.style={rectangle, rounded corners=4pt, minimum width=6.8cm,
                minimum height=0.6cm, text centered, font=\scriptsize\bfseries},
  week/.style={rectangle, rounded corners=3pt, minimum width=1.85cm,
               minimum height=0.8cm, text centered, draw=black!50,
               fill=white, font=\scriptsize},
  outcome/.style={rectangle, rounded corners=3pt, minimum width=1.85cm,
                  minimum height=0.6cm, text centered, draw=black!40,
                  fill=gray!10, font=\scriptsize},
  arr/.style={-{Stealth[length=1.8mm]}, thick}
}

\title{VisionAssist: An Open-Source Smartphone Assistant for AI-Based Visual Accessibility}
\author{Ay\c{s}e \"{O}zlem \c{C}al{\i}\c{s}kan \quad Jordi Sanchez-Riera}
\date{\small Institut de Rob\`otica i Inform\`atica Industrial (CSIC-UPC)}

\newcommand{\cmark}{\textcolor{green!60!black}{\ding{51}}}
\newcommand{\xmark}{\textcolor{red}{\ding{55}}}

\begin{document}
\maketitle

\begin{abstract}
People with low vision often face challenges in performing everyday tasks that require interpreting visual information. We present \textbf{VisionAssist}, an open-source mobile application designed to improve independence by providing AI-powered visual assistance through a smartphone. The application integrates three complementary functionalities within a single interface. First, it enables users to locate specific objects by analyzing the live camera feed. Second, it generates spoken descriptions of captured images, allowing users to identify visual content such as food labels, documents, and everyday objects. Third, it integrates with the smartphone's contacts and calendar to facilitate emergency calls and provide voice-based reminders. The application supports hands-free interaction through voice commands and delivers all feedback using text-to-speech synthesis, making it fully accessible to users with visual impairments. By combining multiple assistive services into a unified platform and releasing the project as open-source software, the proposed solution aims to encourage community contributions and accelerate the development of accessible technologies. The source code is publicly available at: \url{https://github.com/AOzlemC/LowVisionProject.git}. 
\end{abstract}

\section{Introduction}

Visual impairment affects an estimated 285 million people worldwide, substantially limiting their ability to navigate, perceive, and interact with their surroundings independently \cite{who2023}. Everyday activities such as locating misplaced objects, reading printed text, identifying products, or managing personal schedules remain challenging despite the widespread availability of smartphones. Although mobile devices have become increasingly powerful, general-purpose applications do not adequately address the accessibility needs of blind and low-vision (BLV) users. Consequently, there is a growing demand for intelligent assistive technologies that enable natural and independent interaction with the physical and digital environment.

Recent advances in artificial intelligence, particularly the emergence of Vision-Language Models (VLMs) and Multimodal Large Language Models (MLLMs), have transformed the capabilities of mobile assistive systems. By jointly reasoning over visual and textual information and supporting natural-language interaction, these models enable users to obtain scene descriptions, ask follow-up questions, recognize objects, read printed text, and retrieve contextual information from images. Their conversational interface makes them especially well suited for BLV users, allowing visual perception to be replaced by spoken dialogue in a natural and intuitive manner.

At the same time, significant progress in model compression, efficient architectures, and mobile hardware acceleration has made it increasingly feasible to deploy powerful multimodal models on resource-constrained devices such as smartphones. This has led to the rapid development of AI-powered assistive applications that address a variety of individual tasks, including scene description, optical character recognition (OCR), object recognition, navigation assistance, and obstacle detection~\cite{findmythings,lookout,seeingai,envisionai,sullivanplus,blindsquare,navilens,voicedreamscanner}. Other services, such as Be My Eyes~\cite{bemyeyes} or Aira~\cite{aira}, complement AI with remote human assistance, enabling users to obtain reliable answers for complex situations. 

Despite these advances, current assistive solutions remain fragmented. Most applications focus on a single functionality or a limited set of related tasks, requiring users to switch between multiple applications depending on their immediate needs. Furthermore, many state-of-the-art commercial systems rely on proprietary cloud-based models and closed-source software, limiting transparency, extensibility, reproducibility, and personalization. Although recent MLLM-based assistants have significantly improved conversational scene understanding, they generally provide holistic descriptions of the environment rather than supporting targeted, query-driven object localization (e.g., "Where are my keys?"). Similarly, personal productivity functions such as calendar management or communication are typically handled by separate voice assistants, resulting in a fragmented user experience.

To address these limitations, we present a unified, cross-platform assistive mobile application that integrates multimodal visual perception with personal assistant capabilities within a single extensible framework. The proposed system provides three complementary interaction modes. The Find mode enables users to locate specific objects in their surroundings through natural-language queries. The Describe mode supports scene understanding, image description, and text reading using multimodal AI models. Finally, the Personal Assistant mode allows users to perform everyday organizational tasks, including calendar consultation and phone calls, through voice interaction. All interactions are speech-first, allowing users to operate the application hands-free without requiring visual attention to the screen, while responses are delivered through text-to-speech synthesis.

The main contributions of this work are threefold:
\begin{itemize}
    \item A unified mobile assistive framework that combines object localization, scene description, optical character recognition (OCR), contact calling, and calendar assistance into a single speech-driven application.
    \item A carefully designed speech-driven user interface that enables users to interact with the application through voice commands, simplifying the interaction process and improving accessibility.
    \item An open-source, modular implementation that promotes reproducibility, community engagement, and the continued development of new assistive functionalities.
\end{itemize}

\section{Related Work}

Despite the significant advances in Vision--Language Models (VLMs) and, more recently, Multimodal Large Language Models (MLLMs), these technologies cannot yet be directly deployed on resource-constrained mobile devices without important trade-offs. Although recent works have demonstrated remarkable performance in multimodal reasoning and visual understanding~\cite{VLMSurvey,Navisense,NaviGPT,blave,liu2024objectfinder}, their computational and memory requirements often exceed the capabilities of current smartphones. As a result, fully on-device inference typically leads to excessive memory consumption, unstable execution, or long response times, making real-time assistive applications impractical. Consequently, most existing mobile assistive applications rely either on lightweight model variants or on cloud-based inference to provide acceptable performance.

Table~\ref{tab:relatedwork} summarizes the main characteristics of several representative mobile assistive applications. As shown, existing solutions can be broadly categorized according to their primary functionalities and deployment strategies. Some applications provide ´´human assistance´´, connecting users with a remote volunteer or trained agent to interpret visual information. Others are limited to a single mobile platform, supporting either Android or iOS but not both. Applications also differ in their execution model, with some operating entirely on the device while others rely on cloud-based processing. In addition, only a subset of applications offers navigation capabilities.

Compared with these solutions, the proposed application combines multiple AI-powered vision functionalities —including object localization, scene description, and optical character recognition (OCR)— within a unified cross-platform interface. Furthermore, it extends the functionality of conventional visual assistance applications by incorporating voice-controlled personal assistant features, such as contact calling and calendar event retrieval. To the best of our knowledge, the integration of these smartphone assistance capabilities into the same speech-driven application has not been provided by existing mobile assistive solutions.

\begin{table*}[t]
\centering
\scriptsize
\caption{Comparison of existing mobile assistive applications for people with visual impairments. Object denotes object localization, Scene denotes scene description, OCR refers to optical character recognition, Nav. indicates navigation support, and Human represents remote human assistance.}
\label{tab:relatedwork}
\renewcommand{\arraystretch}{1.15}
\begin{tabular}{lccccccccc}
\toprule
\textbf{Application} &
\textbf{Object} &
\textbf{Scene} &
\textbf{OCR} &
\textbf{Nav} &
\textbf{Human} &
\textbf{Calls} &
\textbf{Calendar} &
\textbf{Offline} &
\textbf{Platform} 
\\
\midrule
Aira                  & \cmark & \cmark & \cmark & \cmark & \cmark & \xmark & \xmark & \xmark & iOS, Android \\
Be My Eyes            & \cmark & \cmark & \cmark & Partial & \cmark & \xmark & \xmark & \xmark & iOS, Android \\
Find My Things        & \cmark & \xmark & \xmark & \xmark  & \xmark & \xmark & \xmark & \cmark & iOS \\
Google Lookout        & \cmark & \cmark & \cmark & Partial & \xmark & \xmark & \xmark & Partial & Android \\
Seeing AI             & \cmark & \cmark & \cmark & Partial & \xmark & \xmark & \xmark & Partial & iOS \\
Envision AI           & \cmark & \cmark & \cmark & Partial & \xmark & \xmark & \xmark & Partial & iOS, Android \\
Sullivan+             & \cmark & \cmark & \cmark & Partial & \xmark & \xmark & \xmark & Partial & iOS, Android \\
BlindSquare           & \xmark & \xmark & \xmark & \cmark  & \xmark & \xmark & \xmark & Partial & iOS \\
NaviLens              & Marker & \xmark & Limited & \cmark & \xmark & \xmark & \xmark & \cmark & iOS, Android \\
Voice Dream Scanner   & \xmark & \xmark & \cmark & \xmark  & \xmark & \xmark & \xmark & \cmark & iOS \\
\textbf{VisionAssist (ours)} & \cmark & \cmark & \cmark & \xmark & \xmark & \cmark & \cmark & Partial & iOS, Android \\
\bottomrule
\end{tabular}
\end{table*}

\section{Application Workflow}

\begin{figure}[h]
\centering
\includegraphics[width=0.60\textwidth]{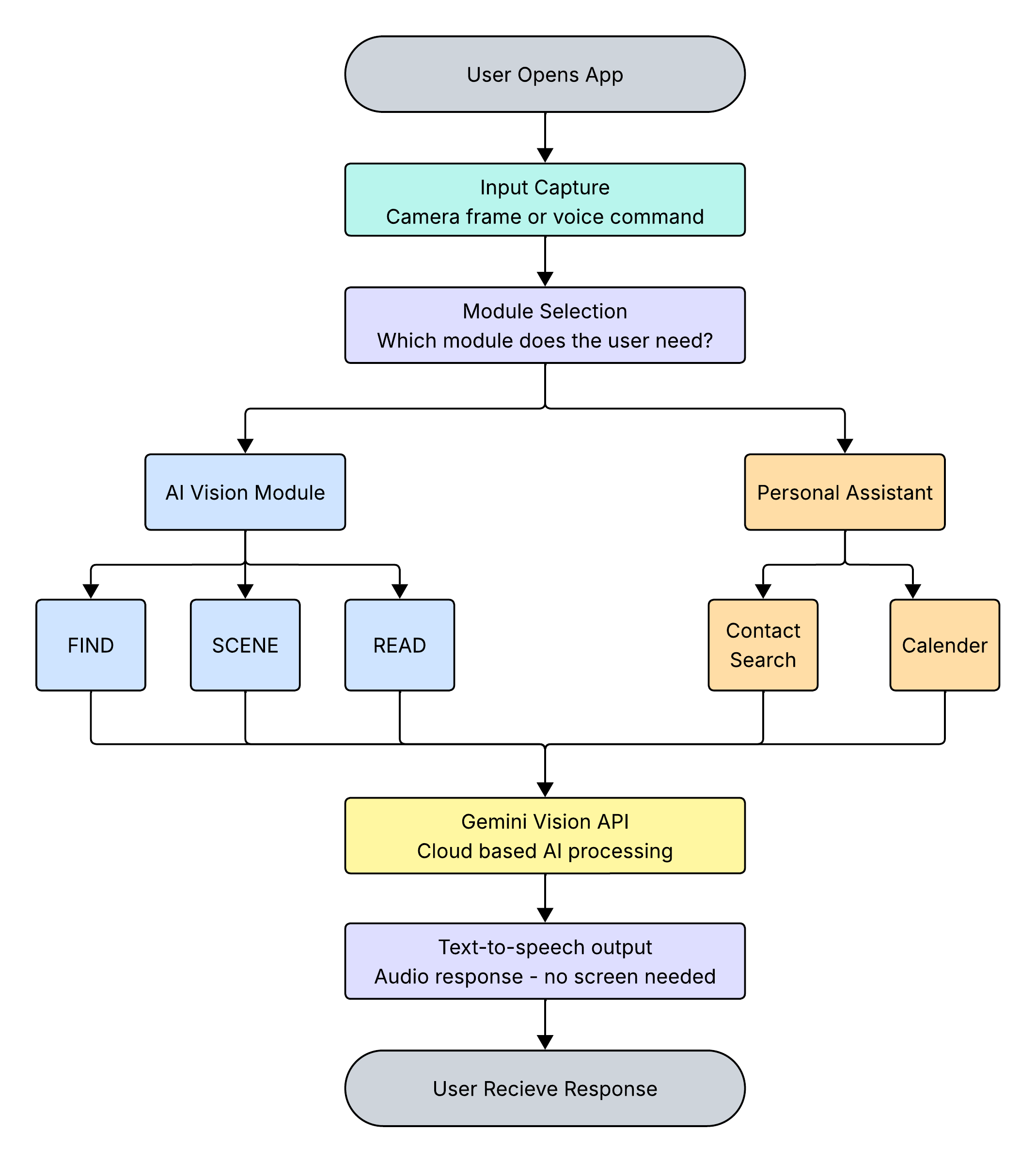}
\caption{General workflow of the mobile intelligent object localization assistant. User input is captured via camera frame or voice command, routed through the AI Vision Module (FIND, SCENE, READ) or Personal Assistant (Contact Search, Calendar), processed by the Gemini Vision API, and returned as Text-to-Speech audio output.}
\label{fig:workflow}
\end{figure}

The application is developed using a modular architecture, allowing it to be easily extended with new functionalities in the future while also facilitating improvements to existing features. The workflow is designed to be intuitive and user-friendly, consisting of a sequence of well-defined stages, as illustrated in Figure~\ref{fig:workflow}.

When the user launches the application, the initial screen captures both an image from the device's camera and the user's voice command. The user is then presented with a set of selectable options. Specifically, the application provides two operating modes: the \textbf{AI Vision} mode and the \textbf{Personal Assistant} mode. Once the desired mode and its corresponding options have been selected, the collected data are passed to a lightweight multimodal large language model (MLLM), which generates a textual response. Finally, the generated text is converted into speech through a text-to-speech (TTS) module, providing the user with an audible response.

\subsection{AI Vision}

The AI Vision module is designed to support three computer vision tasks: \textit{object localization}, \textit{scene description}, and \textit{optical character recognition (OCR)}. For each task, the application captures an image from the smartphone's camera and combines it with a task-specific predefined prompt. The captured image and the corresponding prompt are then transmitted to a cloud-based AI module, which processes the multimodal input and generates an appropriate textual response. Offloading inference to the cloud enables the use of a better Vision--Language Model while avoiding the computational and memory limitations of mobile devices. The generated response is subsequently converted into speech through the text-to-speech (TTS) module, enabling hands-free interaction for the user.

\textbf{Object localization.} In this task, the user specifies a target object via a voice command (e.g., "find my keys"), which a speech recognition module transcribes into text. To ensure system robustness and prevent silent failures, the application immediately validates this input; if no object is specified, it issues a spoken error message prompting the user to try again. Once validated, the transcribed object name and a captured image are forwarded to the AI module with a predefined prompt instructing it to locate the item within two sentences. The model analyzes the scene and generates a concise spatial description of the object's location (e.g., "top-right of the frame"), which a text-to-speech (TTS) module then reads aloud to confirm the result to the user.

\begin{quote}
\textit{Prompt: ``Find [object] in this image. Tell me if it's visible and where it is (e.g., center, top-left, bottom-right). Be brief, max 2 sentences.''}
\end{quote}

\textbf{Scene description.} In this task, the image captured by the smartphone's camera is directly passed to the AI module, which generates a natural language description of the observed scene. Since large language models (LLMs) often produce lengthy responses that may include unnecessary details, the prompt is carefully designed to generate concise and informative descriptions. Specifically, it imposes response length constraints to avoid verbose output, making the generated description more suitable for a listening-based interface. The resulting text is then converted into speech using a text-to-speech (TTS) synthesizer, enabling the user to receive the scene description audibly.

\begin{quote}
\textit{Prompt: ``Describe this scene for a blind person in 2--3 short sentences. Be concise.''}
\end{quote}

\textbf{Optical character recognition (OCR).} In this task, the image captured by the smartphone's camera is expected to contain readable text, such as a product label, street sign, restaurant menu, or any other printed document with recognizable characters. The captured image is then sent to the AI module, which performs optical character recognition (OCR) to extract the text. The recognized text is subsequently converted into speech using a text-to-speech (TTS) module, enabling the user to hear the extracted information.

\begin{quote}
\textit{Prompt: ``Extract and read all text in this image clearly.''}
\end{quote}

\subsection{Personal Assistant}

The Personal Assistant module enables users to access common smartphone functionalities, such as making phone calls and managing their personal agenda, entirely through voice commands. By eliminating the need for manual interaction, the module provides accessible, voice-driven quick actions for frequently performed daily tasks. Each successfully executed action is immediately followed by a text-to-speech (TTS) confirmation, ensuring continuous auditory feedback and maintaining the speech-first interaction paradigm throughout the application. As a result, users can complete all supported tasks without needing to look at or interact with the screen. Unlike the AI Vision module, the personal assistant module works entirely on the device and does not require an internet connection, ensuring reliable access to calendar information even when offline.

\textbf{Phone calls.} In this mode, the user can initiate a phone call by specifying a contact's name through either text or voice input. Voice commands are first transcribed into text and then matched against the device's native contact list using the Expo Contacts API. Once a matching contact is found, the application initiates the phone call automatically. The module supports natural language commands such as \textit{"Call Mom"},  \textit{"Ring John"} and \textit{"Phone [name]"}, providing a flexible and intuitive user experience. Throughout the interaction, the application delivers text-to-speech (TTS) confirmations to inform the user of each step, ensuring continuous auditory feedback.

\textbf{Calendar Events.} In this mode, the assistant retrieves events from the device's native calendar using the Expo Calendar API and presents them to the user through text-to-speech (TTS), ordered chronologically. The module supports natural language queries such as \textit{"Any meetings today?"}, \textit{"Check tomorrow's schedule"} and \textit{"What's happening this week?"}. Based on the user's request, the application automatically infers the appropriate time interval (e.g., the current day, the next day, or the following week) and retrieves the corresponding events.

\section{Implementation and Design}

The application is implemented following the workflow illustrated in Figure~\ref{fig:workflow}. The user interface, including all screens and interactive components, is developed using React Native together with the Expo framework~\cite{expo}. This technology stack was selected for its cross-platform capabilities, enabling the application to run on both Android and iOS devices from a single codebase.

\begin{figure}[h]
\centering
  \subfloat[]{\includegraphics[width=0.248\textwidth]{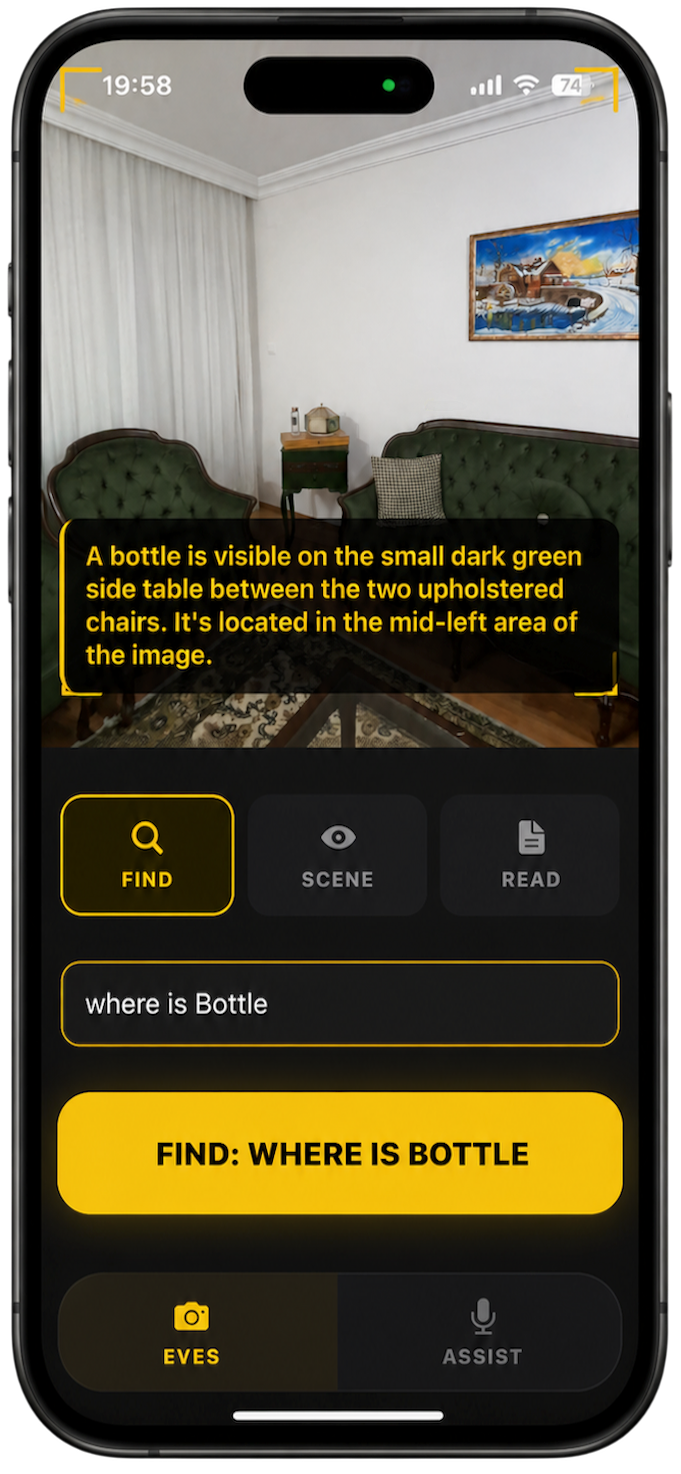}}%
  \subfloat[]{\includegraphics[width=0.248\textwidth]{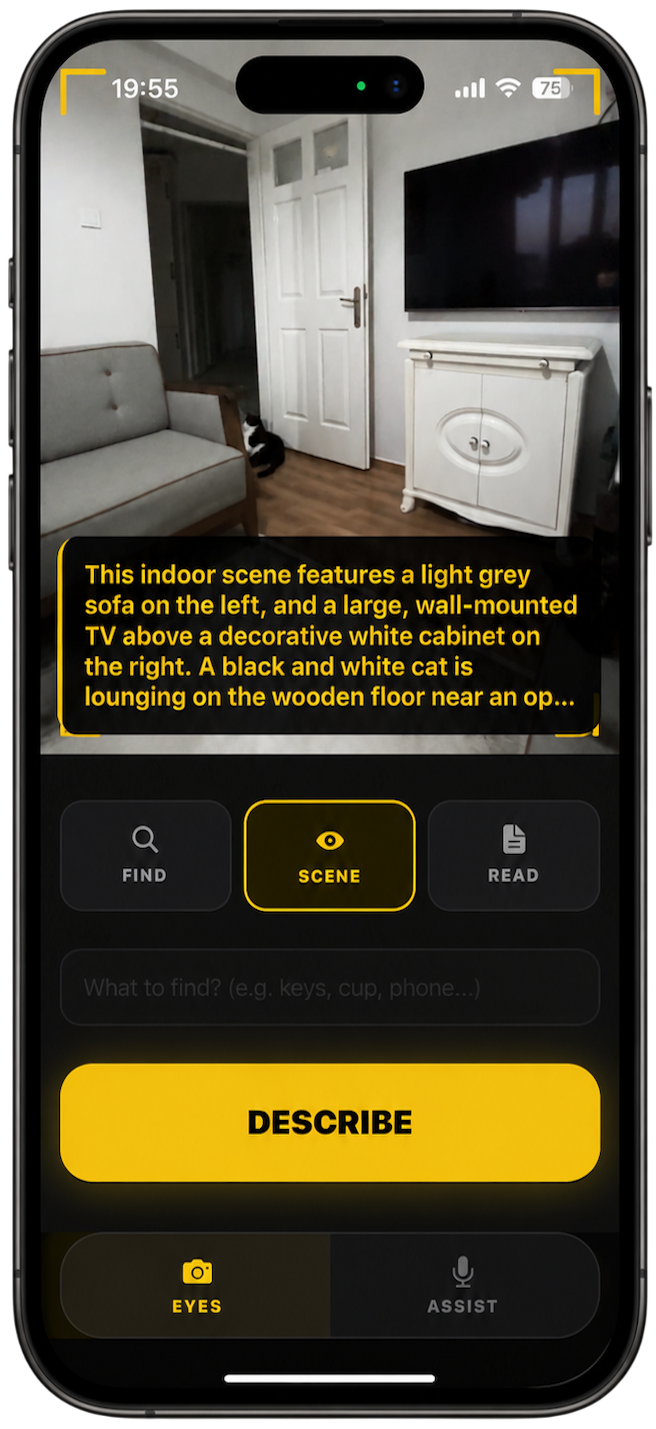}}%
  \subfloat[]{\includegraphics[width=0.248\textwidth]{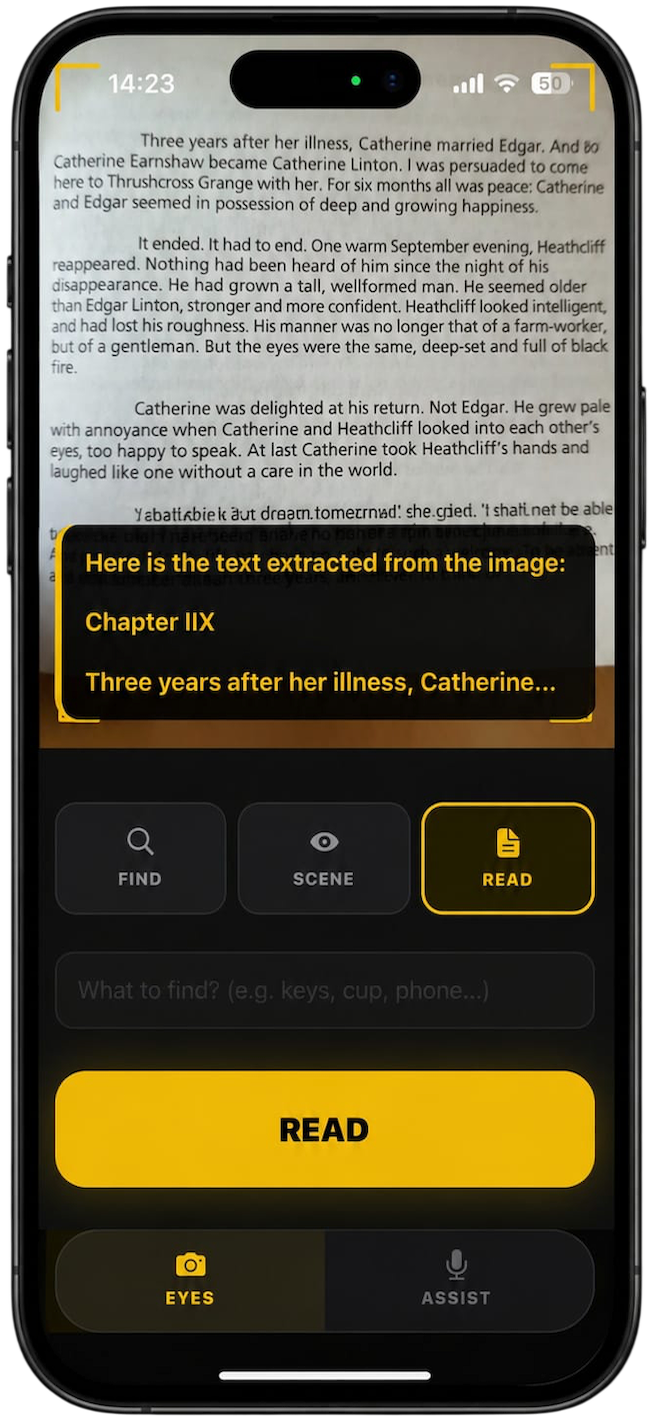}}%
  \subfloat[]{\includegraphics[width=0.248\textwidth]{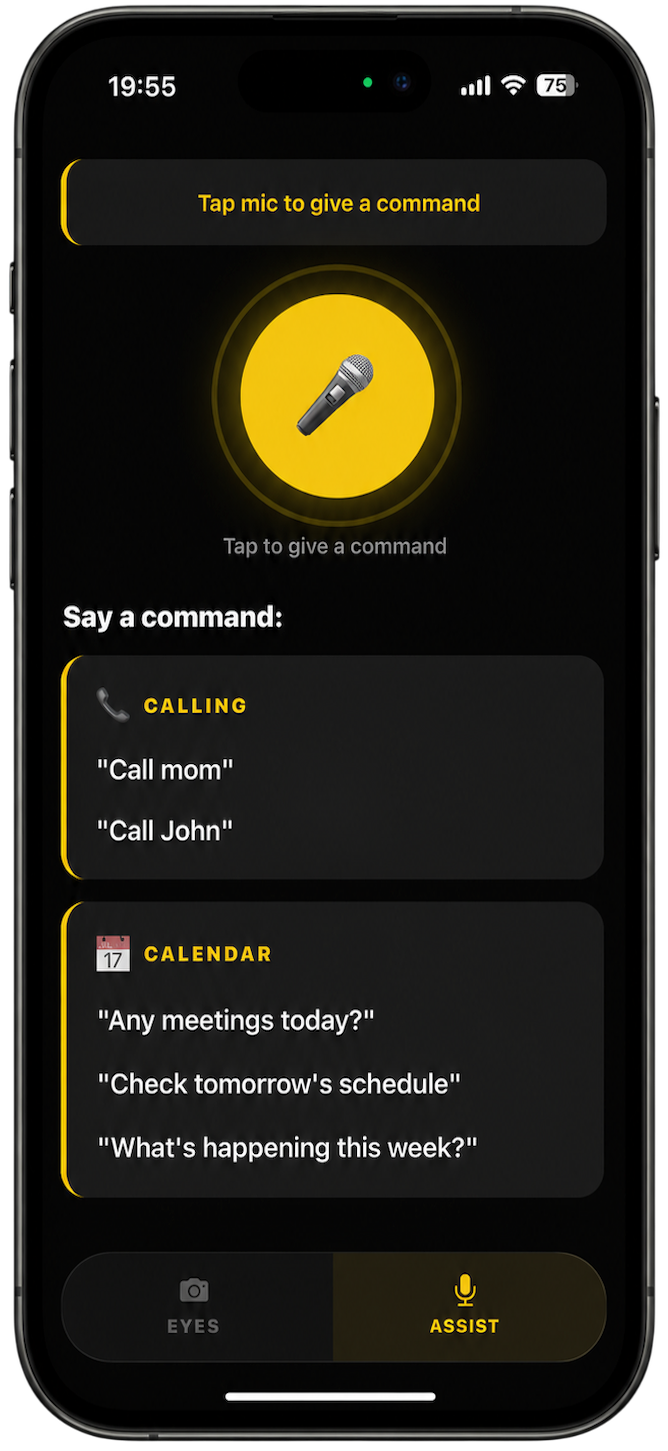}}
\caption{Application interface screenshots. (a) Example of output for the Object recognition mode where the query corresponds to the spoken word ``Bottle''; (b) Example of Scene description mode where camera captures an indoor living room and the model returns a spoken 2--3 sentence description; (c) OCR output on a visible text extracted from the camera frame; (d) Personal Assistant Module presenting the available voice-controlled functionalities, including contact calling and calendar event retrieval.}
\label{fig:finalapp}
\end{figure}

Figure~\ref{fig:finalapp} presents screenshots of the main application interfaces. Figure~\ref{fig:finalapp}(a) shows the home screen, from which the user can choose between the AI Vision module and the Personal Assistant module. Figures~\ref{fig:finalapp}(b)--(d) illustrate the three AI Vision modes (Find, Scene, and Read), while Figure~\ref{fig:finalapp}(e) shows the Personal Assistant interface.

The AI Vision functionalities are implemented using the Gemini Vision API~\cite{gemini}, specifically the \texttt{gemini-2.5-flash} model, which provides multimodal image understanding while maintaining low response latency. The generated textual responses are converted into speech using a text-to-speech (TTS) engine configured with a speech rate of (0.9$\times$) the default speed, a value empirically selected to improve intelligibility and listening comfort for users.


\section{Experiments}
\label{sec:experiments}

Several experiments were conducted to evaluate the performance and effectiveness of the proposed application. First, multiple AI models were compared to identify the most suitable model for the vision-based tasks. After selecting the best-performing model, its performance was evaluated across the three AI Vision functionalities: object localization, scene description, and optical character recognition (OCR). In addition, the functionality and responsiveness of the Personal Assistant module were assessed through a series of task-oriented evaluations. The following sections describe the experimental setup, methodology, and results for each of these evaluations in detail.

\subsection{AI Backbone Selection}

Several state-of-the-art Vision--Language Models (VLMs) were subsequently evaluated for on-device deployment. Qwen2-VL~\cite{qwen} was initially considered because of its strong multimodal capabilities; however, compatibility limitations prevented stable deployment on the target mobile platform. LLaVA~\cite{llava} received the most extensive evaluation due to its promising performance reported in the literature. Although the model successfully completed one or two inference cycles, it consistently crashed under sustained execution because of out-of-memory (OOM) conditions on the 8,GB device, making it unsuitable for continuous operation. Florence~\cite{florence} was also evaluated experimentally but exhibited computational and memory requirements that exceeded the practical capabilities of the target hardware. Finally, Moondream~\cite{moondream}, despite its compact architecture and apparent suitability for edge deployment, repeatedly failed during inference by returning the generic response "I couldn't identify the image," preventing reliable execution.

Overall, the experimental evaluation demonstrated that none of the assessed on-device Vision--Language Models (VLMs) provided the level of stability, reliability, and performance required for the proposed application under the available hardware constraints. Consequently, the Gemini Vision API was selected as the final AI backend. Although this cloud-based solution requires an internet connection, it consistently produced accurate responses with acceptable latency and substantially greater robustness than the evaluated on-device alternatives. Under standard Wi-Fi connectivity, the Gemini API achieved an average response latency of 2--4 seconds, which was considered acceptable for the intended assistive application. Since the system relies exclusively on the pretrained Gemini model for all vision tasks, no task-specific dataset was collected and no additional training or fine-tuning was performed. A quantitative comparison of the evaluated models is presented in Table~\ref{tab:models}.

\begin{table}[h]
\centering
\caption{Comparison of evaluated models for mobile deployment.}
\label{tab:models}
\vspace{0.2cm}
\begin{tabular}{p{3.5cm} p{2.5cm} p{7.8cm}}
\toprule
\textbf{Model} & \textbf{Deployment} & \textbf{Observed Behaviour} \\
\midrule
Qwen2-VL \cite{qwen} & On-device &
  Limited practical success; mobile compatibility constraints. \\
\addlinespace
LLaVA \cite{llava} & On-device &
  1--2 successful inferences; out-of-memory (OOM) crashes under RAM pressure. \\
\addlinespace
Florence \cite{florence} & On-device &
  Insufficient practical performance on target device. \\
\addlinespace
Moondream \cite{moondream} & On-device &
  Repeatedly returned ``I couldn't identify the image''; unreliable
  inference despite compact size. \\
\addlinespace
Gemini \cite{gemini} & Cloud  &
  Accurate and responsive; latency increased over sustained use;
  connectivity-dependent.  \\
\bottomrule
\end{tabular}
\end{table}

\subsection{Application Validation}

The proposed application was validated in two representative indoor environments, namely a home and an office, under typical indoor lighting conditions. Each functionality was evaluated independently using multiple images and user queries representative of real-world usage scenarios. The evaluation was conducted separately for each of the application's operating modes to assess its performance, reliability, and overall functionality under consistent testing conditions.

\subsection{AI Vision}

\textbf{Object Localization} 
\vspace{1mm} \\
\noindent
The object localization functionality was evaluated using 10 commonly encountered household and office objects. For each object, five images were captured from different viewpoints and orientations, resulting in a diverse test set representative of everyday usage conditions. The AI model was assessed based on its ability to correctly identify the requested object and report its approximate location within the scene.

Most objects were detected reliably across the evaluated images. The phone and remote control achieved 100\% detection accuracy, as their distinctive visual characteristics enabled consistent recognition under all tested conditions. The majority of recognition failures occurred in cluttered scenes or under low-light conditions, where object visibility and image quality were reduced. The per-object recognition accuracy is summarized in Table~\ref{tab:find_accuracy}.

\begin{table}[h]
\centering
\caption{Object recognition accuracy per object.}
\label{tab:find_accuracy}
\vspace{0.2cm}
\resizebox{\textwidth}{!}{%
\begin{tabular}{lccccccccccc}
\toprule
 & \textbf{Mouse} & \textbf{Pen} & \textbf{Cup} & \textbf{Keyboard} & \textbf{Bottle} & \textbf{Phone} & \textbf{Keys} & \textbf{Book} & \textbf{Glasses} & \textbf{Remote} & \textbf{Overall} \\
\midrule
Accuracy & 80\% & 80\% & 80\% & 80\% & 80\% & 100\% & 80\% & 80\% & 80\% & 100\% & \textbf{84\%} \\
\bottomrule
\end{tabular}}
\end{table}

\noindent
\textbf{Scene Description}
\vspace{1mm} \\
\noindent
The scene description functionality was evaluated using a dataset of 10 images depicting a variety of everyday indoor scenarios. For each image, the AI model generated a concise description consisting of two to three sentences. Across all test cases, the responses were informative, coherent, and consistently adhered to the prompt-imposed length constraints. This controlled response length made the generated descriptions well suited for direct delivery through the text-to-speech (TTS) module, ensuring a natural and efficient listening experience. An example of a generated scene description is presented below.

\begin{quote}
\textbf{Input:} Living room camera frame (sofa, TV, cabinet, cat).\\
\textbf{Output:} \textit{``This indoor scene features a light grey sofa on the left, and a
large, wall-mounted TV above a decorative white cabinet on the right. A black and white
cat is lounging on the wooden floor near an open door.''}
\end{quote}

\noindent
\textbf{Optical character recognition (OCR)}
\vspace{1mm} \\
\noindent

Similarly, the optical character recognition (OCR) functionality was evaluated using 10 images containing a variety of printed text sources, including product labels, signs, and documents. The AI model successfully extracted the text in 9 out of the 10 test images, achieving an overall accuracy of 90\%. The only failure occurred for an image containing a stylized handwritten label, which proved challenging for the OCR system. An example of a successful OCR result is presented below.

\begin{quote}
\textbf{Input:} Product label (shampoo bottle).\\
\textbf{Output:} \textit{``Head \& Shoulders Classic Clean Shampoo. 400\,ml. For normal hair.
Directions: Apply to wet hair, lather, rinse. Repeat if necessary.''}
\end{quote}

\subsection{Personal Assistant}

\noindent
\textbf{Contact Search}
\vspace{1mm} \\
\noindent

To evaluate the contact access functionality of the Personal Assistant module, we conducted a series of experiments in which users initiated phone calls using natural voice commands. The evaluated queries, listed in Table~\ref{tab:contacts}, included a variety of contact names and natural language expressions. A total of 10 voice queries were tested, and the corresponding results are summarized in Table~\ref{tab:contacts}.

The application successfully matched and retrieved contacts when the spoken query exactly matched or partially corresponded to the stored contact name. Most failures occurred when the voice command was overly informal or did not contain a substring that could be associated with any contact in the device's contact list. Overall, the results demonstrate that the contact retrieval mechanism performs reliably for common voice commands while highlighting the importance of recognizable name matching for successful call initiation.

\begin{table}[h]
\centering
\caption{Contact Search results. Ok = correct contact found and call initiated.}
\label{tab:contacts}
\vspace{0.2cm}
\begin{tabular}{llll}
\toprule
\textbf{Query} & \textbf{Stored Name} & \textbf{Result} & \textbf{Note} \\
\midrule
``Call mom''          & Mom              & Ok              & Exact match \\
``Call dad''          & Dad              & Ok              & Exact match \\
``Call John''         & John Smith       & Ok              & Partial match \\
``Call Sarah''        & Sarah Johnson    & Ok              & Partial match \\
``Call Dr. Ahmed''    & Dr. Ahmed Yilmaz & Ok              & Partial match \\
``Call Emma''         & Emma Clarke      & Ok              & Partial match \\
``Call Alex''         & Alex             & Ok              & Exact match \\
``Call Olivia''       & Olivia Brown     & Ok              & Partial match \\
``Call Lisa''         & Lisa Taylor      & \texttimes{}    & No match found \\
``Call the doctor''   & Dr. Ahmed Yilmaz & \texttimes{}    & Descriptive query unsupported \\
\midrule
\textbf{Accuracy}     &                  & \textbf{8/10 (80\%)} & \\
\bottomrule
\end{tabular}
\end{table}

\noindent
\textbf{Calendar Events}
\vspace{1mm} \\
\noindent

Similarly, the calendar event retrieval functionality was evaluated using 10 representative voice queries covering events scheduled for the current day, the following day, and the upcoming week. The corresponding queries and results are summarized in Table~\ref{tab:calendar}.

The Personal Assistant module achieved 100\% accuracy across all test cases. Since this functionality operates entirely on the device through the native Expo Calendar API, its performance is independent of AI model variability and network connectivity. For every query, the application successfully retrieved the relevant events and presented them in chronological order, correctly reporting the event title, date, and time through the text-to-speech (TTS) module. These results demonstrate the reliability of the on-device calendar integration for voice-based personal assistance.

\begin{table}[h]
\centering
\caption{Calendar Events results. Ok = correct events retrieved and read aloud.}
\label{tab:calendar}
\vspace{0.2cm}
\begin{tabular}{llll}
\toprule
\textbf{Query} & \textbf{Look-ahead} & \textbf{Events} & \textbf{Result} \\
\midrule
``Any meetings today?''          & 1 day  & 2 & Ok \\
``Check today's schedule''       & 1 day  & 0 & Ok (correctly reported none) \\
``Check tomorrow's schedule''    & 2 days & 1 & Ok \\
``Any events tomorrow?''         & 2 days & 3 & Ok \\
``What's happening this week?''  & 7 days & 5 & Ok \\
``Any meetings this week?''      & 7 days & 5 & Ok \\
``Do I have anything today?''    & 1 day  & 1 & Ok \\
``What's on my calendar?''       & 1 day  & 2 & Ok \\
``Check my schedule''            & 1 day  & 0 & Ok (correctly reported none) \\
``Upcoming events''              & 7 days & 4 & Ok \\
\midrule
\textbf{Accuracy} & & & \textbf{10/10 (100\%)} \\
\bottomrule
\end{tabular}
\end{table}

\section{Conclusions}

This work presents a mobile assistive application that combines AI-powered vision capabilities with voice-based smartphone interaction to address key limitations of existing visual assistance tools. A systematic evaluation of several on-device Vision--Language Models (VLMs) highlighted the challenges of deploying multimodal AI on resource-constrained devices and motivated the adoption of a cloud-based vision pipeline. The resulting hybrid architecture provides a practical balance between performance, reliability, and accessibility.

The final application offers five core functionalities: object localization, scene description, optical character recognition (OCR), contact search, and calendar event retrieval. All features are accessible through a speech-first interface, enabling users to perform tasks without relying on visual interaction with the screen. Moreover, the cross-platform implementation based on React Native and the Expo framework provides a modular and extensible foundation for future research in mobile accessibility.

The main limitation of the proposed system is its reliance on an internet connection for the AI Vision module. Future work will focus on integrating lightweight on-device Vision--Language Models as mobile AI technology matures, conducting user studies with visually impaired participants, and performing a more comprehensive quantitative evaluation of the system's performance.

\bibliographystyle{alpha}
\bibliography{sample}

\end{document}